\documentclass[10pt, a4paper]{article}

\usepackage[final]{lrec-coling2024} % this is the new style

\usepackage{rotating}
\usepackage{colortbl}
\usepackage{longtable}
\usepackage{booktabs}
\usepackage{graphicx}
\usepackage{amssymb}

\title{GPTEval: A Survey on Assessments of ChatGPT and GPT-4}

\name{Rui Mao$^{\spadesuit}$, Guanyi Chen$^{\clubsuit}$, Xulang Zhang$^{\spadesuit}$, Frank Guerin$^{\blacklozenge}$, and Erik Cambria$^{\spadesuit}$} 

\address{$^{\spadesuit}$Nanyang Technological University, Singapore \\
         $^{\clubsuit}$Central China Normal University, China \\
         $^{\blacklozenge}$University of Surrey, United Kingdom\\
         rui.mao@ntu.edu.sg; g.chen@ccnu.edu.cn; xulang001@e.ntu.edu.sg;\\f.guerin@surrey.ac.uk; cambria@ntu.edu.sg\\}

\abstract{
The emergence of ChatGPT has generated much speculation in the press about its potential to disrupt social and economic systems. Its astonishing language ability has aroused strong curiosity among scholars about its performance in different domains. There have been many studies evaluating the ability of ChatGPT and GPT-4 in different tasks and disciplines. However, a comprehensive review summarizing the collective assessment findings is lacking. The objective of this survey is to thoroughly analyze prior assessments of ChatGPT and GPT-4, focusing on its language and reasoning abilities, scientific knowledge, and ethical considerations. Furthermore, an examination of the existing evaluation methods is conducted, offering several recommendations for future research.
 \\ \newline  }

\begin{document}

\maketitleabstract

\section{Introduction}

ChatGPT~\citep{openai2023chatgpt} has generated significant scholarly interest across various disciplines due to its impressive dialogue-based task-processing capabilities. This has enabled users to explore and evaluate its performance across a wide range of tasks and disciplines, thereby sparking considerable enthusiasm in the field of Artificial Intelligence (AI). While many researchers have concentrated on evaluating ChatGPT and GPT-4~\citep{openai2023gpt4} within their specific domains of expertise, a comprehensive review encompassing the assessments in multiple tasks and disciplines can offer a holistic understanding of the strengths and limitations of these GPT models. We focus on ChatGPT and GPT-4, because they are state-of-the-art (SOTA) large language models (LLMs). The scope of our survey encompasses quantitative evaluations carried out on ChatGPT or GPT-4, specifically focusing on their language proficiency, scientific knowledge, and ethical considerations. Our main findings are summarized as follows:\\
\noindent \textbf{\textit{a}}) ChatGPT and GPT-4 are strong in language understanding and generation, adeptly engaging in user interactions through dialogues, enabling them to tackle diverse NLP tasks and provide explanatory outputs. However, their current status falls short of being a comprehensive AI, as their performance lags behind expert models in numerous domains involving domain-specific knowledge.

\noindent \textbf{\textit{b}}) ChatGPT performs satisfactorily in general science knowledge and can answer science questions that desire open responses. However, it can also make mistakes, especially for questions that require multi-step reasoning.
The exceptional language proficiency poses challenges for users in assessing the accuracy of factual information, giving rise to a range of ethical concerns.

\noindent \textbf{\textit{c}}) Existing evaluation methods may be unreliable. The current evaluation methods heavily depend on prompt engineering and benchmark datasets. Varying prompts can yield disparate evaluation results. Additionally, the comparison of expert systems often relies on (in-domain) datasets that were utilized for training those systems. It remains uncertain if the examined data, such as public datasets and scientific knowledge, have been inadvertently exposed during the training of ChatGPT and GPT-4. These factors may contribute to an unfair comparison between LLMs and their respective baselines.

The contributions of this work are threefold: (1)~We conduct a comprehensive survey of recent assessments focusing on the language proficiency and scientific knowledge of ChatGPT and GPT-4. (2) We compare their assessment results across various tasks and disciplines to highlight the strengths and weaknesses of the GPT models. (3) We critically analyze the existing assessment methods employed, offering recommendations for future evaluation studies and delivering our ethical considerations associated with the GPT models.

\section{Language and Reasoning Ability}

\subsection{Classic NLP Tasks}

\noindent \textbf{Dialogue.} 
\citet{cabrera2023zeno} compared several chatbots with a novel LLM evaluation toolkit, termed Zeno Build. The performance was evaluated by Critique metrics\footnote{\url{https://docs.inspiredco.ai/critique/}} such as ChrF~\citep[a character- and word n-grams similarity-based metric,][]{popovic2015chrf}, BERTScore~\citep[a BERT embedding similarity-based metric,][]{zhangbertscore}, and UniEval Coherence~\citep[a coherence probability-based metric,][]{zhong2022towards}. They found that ChatGPT surpassed all the baselines, e.g., GPT-2~\citep{radford2019language}, LLaMa~\citep{touvron2023llama}, Alpaca~\citep{taori2023alpaca}, Vicuna~\citep{vicuna2023}, MPT-Chat~\citep{MosaicML2023Introducing}, and Cohere Command\footnote{\url{https://docs.cohere.com/docs/command-beta}} in the three evaluation metrics. However,~\citet{cabrera2023zeno} also highlighted that ChatGPT exhibited vulnerabilities that were noticeable in various aspects, such as occurrence of hallucinations, inadequate exploration for additional information, and repetition of content. Although ChatGPT exceeded other LLMs,~\citet{bang2023multitask} observed that SOTA models still outperformed ChatGPT on task-oriented dialogue, and open-domain knowledge-grounded dialogue, based on automatic evaluation metrics. 

\noindent \textbf{Generation} was almost evaluated on text-to-text generation tasks. For machine translation (MT), an early assessment suggested SOTA MT systems could defeat ChatGPT by a large margin~\citep{bang2023multitask}. Later on, by testing on more language pairs, more datasets, and better prompts,~\citet{hendy2023good,jiao2023chatgpt} found ChatGPT yielded competitive performance for high-resource languages, but still had limited capabilities for low-recourse languages. Through a large-scale human evaluation and error analysis by expert translators,~\citet{karpinska2023large} found that, when doing paragraph-level translation, ChatGPT's translations were overwhelmingly preferred compared to those from Google Translate and it largely reduced errors, including mistranslation, grammatical errors, inconsistency errors, and more. 

For summarization, by testing on multiple summarization datasets in multiple languages,~\citet{qin2023chatgpt,bang2023multitask,lai2023chatgpt} observed that ChatGPT largely underperformed SOTA systems in doing either abstractive or extractive summarization.~\citet{zhang2023extractive} found that its performance, especially the faithfulness, improved if we asked ChatGPT to first extract salient sentences and then generate the summarization, based on the extracted sentences, although it still lost to SOTA. However, similar to what happened when assessing ChatGPT's translation ability,  This conclusion should be further confirmed by larger populations and rigorous experiments. Later on, a small human evaluation found that annotators could not distinguish summaries generated by ChatGPT from those by humans~\citep{soni2023comparing}, which, however, was then falsified by a much larger manual assessment study~\citep{pu2023summarization}.

A few assessments on data-to-text generation suggested that both ChatGPT and GPT-4 under-performed to fine-tuned T5 and BART~\citep{ren2023you,yuan2023evaluating}, which was probably due to the complexities in expressing relations between data and text in prompts. Although ChatGPT is often considered to perform well in generation tasks that need creativity, there have not been many related assessments. \citet{jentzsch2023chatgpt} asked ChatGPT to produce jokes and suggested that while ChatGPT generated jokes, it struggled to produce ``new'' jokes. Approximately 90\% of the generated jokes were repetitions of the same 25 jokes. \citet{chu_liu_2023} conducted reader experiments and found that ChatGPT wrote more engaging and persuasive short stories than its human counterparts. However, the conclusion was the opposite if the aim was long stories. They suggested that ChatGPT might focus on retaining information from the instruction when writing a long story, which limited its flexibility. 

\noindent \textbf{Affective Computing.} \citet{amin2023wide} compared ChatGPT and GPT-4 to supervised learning models on 13 affective computing domains. The baselines were trained with task-specific datasets using different embedding representations. They found that the RoBERTa embedding-based model exceeded ChatGPT and GPT-4 on 22 out of 34 evaluation tasks. \citet{qin2023chatgpt,bang2023multitask,kocon2023chatgpt} also found that fine-tuned models exceeded ChatGPT on English sentiment analysis and emotion detection tasks.

\noindent \textbf{Information Retrieval.} \citet{wei2023zero} examined ChatGPT on relation extraction, named entity recognition (NER), and event extraction tasks, showing that the performance of the basic version of ChatGPT was much weaker than supervised methods, while largely exceeding 50 shot or fewer shot baselines. Then, they introduced a multi-turn question answering (QA) framework, where the modified querying process did not help ChatGPT to exceed the supervised baseline. \citet{qin2023chatgpt} observed that ChatGPT yielded much weaker performance than fine-tuning-based models on CoNLL03 NER dataset~\citep{sang2003introduction}. \citet{sun2023chatgpt} evaluated ChatGPT and GPT-4 on passage re-ranking tasks, showing that GPT-4 outperformed SOTA supervised baselines across all three datasets, including a multilingual dataset with 10 languages. ChatGPT slightly fell behind SOTA methods, while largely exceeding BM25. \citet{bubeck2023sparks} found that GPT-4 (77.4\% accuracy) significantly outperformed the SOTA model~\citep[40.8\%,][]{payne2020privacy} on a personally identifiable information detection task.

\noindent \textbf{GPT as Human Annotator.} \citet{wang2023chatgpt} compared ChatGPT with SOTA natural language generation evaluation metrics, e.g., BERTScore, BARTScore~\citep{yuan2021bartscore}, ROUGE~\citep{lin2004rouge}, and more. ChatGPT slightly outperformed the strongest BARTScore-based setup on the SummEval dataset~\citep{fabbri2021summeval} in coherence, relevance, consistency, and fluency dimensions, while was surpassed by BARTScore on the NewsRoom dataset~\citep{grusky2018newsroom}. ROUGE-1 yielded the highest correlation scores on the sample- and dataset-level evaluations on the RealSumm dataset~\citep{bhandari2020re}. The above three datasets were used for evaluating text summarization, although presenting inconsistent results. ChatGPT achieved the largest improvements on the OpenMEVA-ROC~\citep{guan2021openmeva} story generation dataset. ChatGPT and baselines were comparable on the BAGEL dataset~\citep{mairesse2010phrase} for informativeness, naturalness, and quality evaluations. \citet{liu2023gpteval} found that GPT-4 could exceed other metrics on the dialogue generation dataset from~\citet{mehri2020usr}. \citet{kocmi2023large} evaluated ChatGPT-generated machine translation accuracy scores with its former versions and other SOTA scoring systems, finding that ChatGPT was less accurate than those baselines. \citet{gilardi2023chatgpt} observed that ChatGPT achieved higher intercoder agreements than MTurk crowd-workers and educated annotators, and maintained the highest annotation accuracy in tweet frame and stance annotation tasks. The relevance annotation accuracy of ChatGPT was comparable to humans, while topic annotations were not as useful as MTurk.

\subsection{Multilingualism} \label{sec:multiling}

\noindent \textbf{Chinese Linguistic Test.} SuperCLUE benchmarks (Xu et al.,~\citeyear{SuperCLUE}) compared ChatGPT-like foundation models in regard to their basic language ability, professional ability, and Chinese-featured ability. By 18th May 2023, they reported that GPT-4 and ChatGPT achieved the second (76.67) and third best (66.18) results after humans (96.50), exceeding other LLMs. GPT-4 and ChatGPT had better basic language ability than the other two metrics in Chinese. Both models achieved human-like accuracy on role-playing, chit-chatting, and coding. However, their ability to understand Chinese poetry, literature, classical Chinese, and couplets was far inferior to that of humans. \citet{huang2023c} also ranked GPT-4 and ChatGPT as top-2 on a multi-discipline (52 subjects) Chinese evaluation.

\noindent \textbf{Multilingual NLP Tasks} were examined by~\citet{lai2023chatgpt}, e.g., multilingual part-of-speech (PoS) tagging, NER, relation classification, NLI, QA, commonsense reasoning, and summarization. The researchers analyzed 36 languages and discovered that task-specific fine-tuned models outperformed ChatGPT in the majority of examined tasks, except PoS tagging. ChatGPT exhibited superior performance in English tasks compared to tasks in other languages; for low- and extremely low-resource languages, ChatGPT performed significantly worse than baselines. Noticeably, despite the use of non-English languages in the target tasks, ChatGPT improved its performance with English prompts. \citet{wei2023zero} found that direct usage of ChatGPT yielded unsatisfying results in Chinese information extraction.  \citet{wang2023cross} tested ChatGPT and GPT-4 on English-to-Chinese and English-to-German summarization, showing that although ChatGPT and GPT-4 exceeded other LLM baselines on a zero-shot setup, they fell behind a fine-tuned mBART-50~\citep{tang2021multilingual} on most of the examined datasets. \citet{bang2023multitask} argued that ChatGPT generally yielded weak performance on low-resource languages in language understanding and generation, while achieving higher proficiency in comprehending non-Latin scripts compared to its proficiency in generating them.

\subsection{Reasoning}

\noindent \textbf{Logical Reasoning} was tested by~\citet{bang2023multitask}. They found 56 out of 60 answers correct (with appropriate prompts) for deductive reasoning, i.e. applying general rules to specific situations or cases. This was stronger than other types of reasoning. 26 out of 30 were scored for abductive reasoning, i.e. forming plausible explanations or hypotheses, based on limited evidence or incomplete information.  33 out of 60 were scored for inductive reasoning, i.e.,  drawing generalized conclusions from examples or specific observations.

\noindent \textbf{Commonsense Reasoning.} \citet{bang2023multitask}  tested ChatGPT via three commonsense datasets, showing that 80 out of 90 of ChatGPT's predictions were correct. ChatGPT was able to give good explanations of the reasoning steps to support its answer. However,~\citet{qin2023chatgpt,laskar2023systematic} showed that the commonsense reasoning accuracy of ChatGPT fell behind fine-tuned baselines. \citet{davis2023benchmarks} found significant flaws in common benchmarks for common sense, including the CommonsenseQA dataset used by~\citet{bang2023multitask}, which he explicitly addressed. \citet{davis2023benchmarks} listed several examples of commonsense and particularly physical reasoning failures that had been found shortly after the release of ChatGPT, and pointed to others. However, there does not exist a thorough assessment of the GPT models' commonsense reasoning ability. More generally,~\citet{davis2023benchmarks}  pointed out that ``many important aspects of commonsense reasoning and commonsense knowledge are not tested in existing benchmark''.~\citet{bubeck2023sparks} probed a small number of their own real-world physical reasoning tasks with GPT-4, finding that it had good knowledge and concluded it was able to learn an understanding of the real-world environment.

\noindent \textbf {Causal Reasoning.} \citet{bang2023multitask} found that 24 out of 30 causes or effects could be correctly identified. \citet{gao2023chatgpt} systematically evaluated event causality identification, causal discovery, and causal explanation generation. Compared to SOTA models, ChatGPT and GPT-4 yielded lower scores in causality identification. They outperformed baseline models on the causal discovery, although the compared models, e.g., BERT-~\citep{devlin2019bert} and RoBERTa-base~\citep{liu2019roberta} were relatively weak. The generation of causal explanations yielded inconsistent findings in terms of AVG-BLEU and ROUGE-l metrics, while the human evaluation affirmed that both GPT models attained a level of accuracy comparable to that of human performance. \citet{kiciman2023causal} examined ChatGPT and GPT-4 on the causal discovery, counterfactual reasoning, and actual causality inferring, finding that they outperformed other LLMs and SOTA models largely on the first two tasks.

\noindent \textbf{Psychological Reasoning} is the ability of humans to reason about other's unobservable mental states (a.k.a Theory of Mind (ToM)). \citet{kosinski2023theory} and~\citet{moghaddam2023boosting} designed sets of False-Belief questions and quantified results suggested that both ChatGPT and GPT-4 had ToM ability, but that was still inferior to a human's. However,~\citet{davmarc2023} pointed out flaws in the~\citet{kosinski2023theory} study because the test material was in the training data. \citet{holterman2023does} tested GPT-3 and GPT-4 on more ToM tasks summarized in~\citet{daniel2000thinking}. They acknowledged the potential problem of the test material being in the training data, so they substituted various nouns in the scenario. However, this is unlikely to be adequate to stop a neural model from generalising from those examples. \citet{borji2023categorical} found that chatGPT failed on a variant of a classic `Sally-Anne Test' (used to test children). \citet{bubeck2023sparks} found that chatGPT answered correctly on a similar variant Sally-Anne, and they further tested on a range of more advanced ToM scenarios, with probing questions, e.g. to infer the counterfactual impact of actions on mental states. They found that GPT-4 had superior abilities and suggested that GPT-4 had a very advanced level of ToM.

\noindent \textbf{Task-Oriented Reasoning.} \citet{qin2023chatgpt} evaluated dialogue, logical reasoning, complex yes/no QA, symbolic reasoning (last letter concatenation and coin flip), date understanding-, and tracking shuffled objects-oriented logical reasoning. However, ChatGPT underperformed fine-tuned baselines on most of the tasks, excluding the logical reasoning tasks. To ascertain whether ChatGPT relies on profound comprehension of truth and logic in their reasoning or merely exploits shallow memorized patterns,~\citet{wang2023can} proposed a dialectical evaluation task, finding that despite displaying high confidence, ChatGPT demonstrated an inability to hold its belief in the truth in a wide range of reasoning tasks, e.g., mathematics, first-order logic, commonsense, and generic reasoning.

\noindent \textbf{Natural Language Inference (NLI)} aims to examine if a statement can be inferred, contradicted, or neutral, compared to another statement. \citet{liu2023evaluating} compared ChatGPT and GPT-4 with RoBERTa. However, both models encountered difficulties when dealing with novel and out-of-distribution data. They yielded relatively modest performance on NLI that needed logical reasoning. \citet{qin2023chatgpt} also proved that the NLI ability of ChatGPT was lower than that of supervised models. Ambiguity is one of the difficulties of NLI. For example, whether ``\emph{John and Anna are not a couple}''  contradicts ``\emph{John and Anna are married}'' depends on whether ``\emph{married}'' means ``\emph{both married}'' or ``\emph{married to each other}''. Given an ambiguous NLI premise,~\citet{liu2023were} asked ChatGPT and GPT-4 to either generate disambiguations of a premise with respect to the hypothesis or recognize disambiguation (i.e., deciding whether the disambiguation is an interpretation of an ambiguous premise). Their human evaluation showed that for the first task, GPT-4 achieved correctness at 32\%, while, for the second task it was at the level of random guessing, suggesting that resolving tricky ambiguity remained challenging for ChatGPT.

\section{Scientific Knowledge}

\subsection{Formal Science}

\noindent \textbf{Mathematics.}~\citet{frieder2023mathematical}  proposed a mathematical benchmark at the graduate level to test ChatGPT's mathematical reasoning, related to textbook exercises, Olympiad problems, proof completion, algebra and probability theory problem solving, and theorem-proof and definition understanding. They found that ChatGPT only achieved a passing grade (50\% of points) on 6 out of 17 testing sets. ChatGPT performed badly on mathematical problem solving, e.g., Olympiad problems, textbook exercises, algebra, and probability theory, while presenting comparatively better grades in definition understanding and proof completion. It seems ChatGPT is not good at solving mathematical problems that are weakly related to language memory or generation, because compared to the textual understanding, e.g., theorem-proofs and definitions, the other takes depend on multi-step reasoning. \citet{qin2023chatgpt} evaluated ChatGPT in arithmetic reasoning, finding that ChatGPT achieved the highest score on 3 out of 6 datasets, while fine-tuning-based methods exceeded ChatGPT on the rest of the 3 datasets. \citet{bubeck2023sparks} showed that GPT-4 largely exceeded the Minerva expert model~\citep{lewkowyczsolving}, while ChatGPT fell behind Minerva on average. From a user interaction perspective,~\citet{Collins2023EvaluatingLM} evaluated the mathematical capabilities of InstructGPT, ChatGPT and GPT-4, and identified several weaknesses, including their performance in algebraic manipulations, tendency towards verbosity, and reliance on memorized solutions.

\noindent \textbf{Computer Science.}~\citet{bordt2023chatgpt} developed an exam with 10 different exercises. ChatGPT and GPT-4 achieved 20.5 and 24 points out of 40 full points, respectively. GPT-4 slightly exceeded the average score (23.9) of 200 students. \citet{bordt2023chatgpt} believed that passing the exam by ChatGPT should not be misconstrued as an indication of its comprehension of computer science, because numerous topics addressed in the exam were extensively documented and readily available online. The coding skills of ChatGPT and GPT-4 largely exceeded the SOTA baselines~\citep{bubeck2023sparks}. GPT-4 even achieved human-level accuracy in LeetCode tests, while ChatGPT yielded lower accuracy, reaching only half of the average human performance. \citet{li2023can} compared in-context learning (ICL)-based ChatGPT to SOTA models on Text-to-SQL tasks. To improve the performance of the examined ICL models, Chain-Of-Thought (COT) and extra knowledge evidence sentences were also incorporated. ChatGPT (40.08\%) exceeded the strongest baseline Codex~\citep[36.47\%,][]{chen2021evaluating}, while largely lagging behind humans (92.96\%) in execution accuracy. 

\subsection{Natural Science}

\noindent \textbf{Physics.} ChatGPT was evaluated as if it is a college student who needs to finish homework, clicker questions, programming exercises, and exams in the first-year Calculus-based Physics~\citep{kortemeyer2023could}. Overall, ChatGPT achieved 53.05\% after weighing different testing modules. This score met the minimum requirement for course credit, yet it adversely affected the overall grade-point average, falling below the necessary threshold for graduation. ChatGPT showed outstanding performance in clicker and programming questions, achieving scores higher than 90\%. However, its performance in homework and exams was subpar. Additionally, ChatGPT's mathematical difficulties in the field of physics lowered its overall score.

\noindent \textbf{Chemistry.}~\citet{clark2023investigating} asked ChatGPT to finish two real chemistry exams with closed- and open-response questions. 44\% closed-response questions were correctly answered by ChatGPT, although this is lower than the student average score (69\%). Conversely, when it came to open-response questions, ChatGPT's performance was even lower than that of the least successful student.

\noindent \textbf{Medicine.}~\citet{gilson2023does,kung2023performance} showed that ChatGPT achieved college student level on the United States Medical Licensing Examination (USMLE). \citet{antaki2023evaluating} found ChatGPT yielded low accuracy on neuro-ophthalmology and high accuracy on general medicine, indicating it had not grasped specialized medical knowledge well. \citet{hirosawa2023diagnostic} examined ChatGPT's diagnosis with 30 clinical vignettes, showing that the rate of top-10 differential suggestions covering the correct diagnosis reached 93.3\%, while the accuracy of top-1 suggestions was just 53.3\% (physicians were 93.3\%). \citet{rao2023assessing} suggested that ChatGPT achieved an impressive final diagnosis accuracy (76.9\%), while its initial diagnosis accuracy was just 60.3\%. \citet{mehnen2023chatgpt} analyzed ChatGPT and GPT-4 diagnoses by 40 common and 10 rare clinical vignettes. They found that GPT-4 achieved 100\% accuracy on the common cases within three possible diagnostic suggestions. ChatGPT reached about 95\% with the same number of suggested diagnoses.

\subsection{Social Science}

\noindent \textbf{Education.} \citet{dahlkemper2023physics} aimed to investigate the extent to which students possessed accurate perception regarding the scientific accuracy and the linguistic quality of ChatGPT's responses. They provided 102 physics students with ChatGPT answers and (masked) expert answers to evaluate the scientific accuracy and linguistic quality of both parties, perceived by the students, based on three physics questions. Despite the fact that all responses generated by ChatGPT in their study were incorrect, imperfect, or misleading, the evaluation results indicated that when confronted with a difficult question, students perceived ChatGPT's scientific accuracy to be on par with that of the expert solution, attributing this to the higher linguistic quality of ChatGPT. However, in cases where the questions were relatively easy, both the scientific accuracy and linguistic quality of the expert's answers was perceived to surpass ChatGPT.

\noindent \textbf{Law.} ChatGPT was tested with four law class exams by~\citet{choi2023chatgpt}, including Constitutional Law, Employee Benefits, Taxation, and Torts. ChatGPT performed better on 12 essay questions than on the 95 multiple-choice questions. Although it passed all four exams, it ranked almost the lowest among law school students in each class. \citet{liu2023evaluating} examined ChatGPT and GPT-4 in Law School Admission Test (LSAT) and the Chinese Civil Service Examination (CCSE). Compared with RoBERTa, the advantage of GPT-4 was much larger than that of ChatGPT. GPT-4 even exceeded the average level of a human on LSAT. Nevertheless, when compared to the human ceiling, significant disparities persisted in both models.

\noindent \textbf{Economics.} ChatGPT was examined with the Test of Understanding of College Economics~\citep[TUCE,][]{geerling2023chatgpt}. It demonstrated proficiency by providing accurate responses to 19/30 microeconomics questions and 26/30 macroeconomics questions. Such performance placed it in the upper echelons, ranking within the top 9\% and top 1\% among 3,255 and 2,789 college students who had successfully finished a full semester of microeconomics and macroeconomics. \citet{xie2023wall} tested ChatGPT in stock predictions. ChatGPT showed weaker performance than traditional machine learning algorithms in numerical feature-based predictions. Through error analysis, it was determined that ChatGPT was also weak in comprehending investor sentiment expressed in text.

\section{Ethical Considerations}

\noindent \textbf{Fairness} asks models to be fair in terms of gender, race, language, culture, and more.  \citet{seghier2023chatgpt,yong2023prompting} found that ChatGPT's responses were much worse in languages other than English (see \S~\ref{sec:multiling} ). \citet{zhuo2023red} assessed two types of social biases, namely gender bias and race bias. They concluded that, for text generation, although biases still existed, ChatGPT had mitigated them to a large extent compared to its predecessors, and that, for dialogue generation, ChatGPT could generate unbiased responses.

\noindent \textbf{Robustness} requires models to maintain their performance when the inputs are different from the training data. Such inputs could be noisy data, outliers and attacks. For noisy data,~\citet{zhuo2023red} tried two NLP datasets while~\citet{ye2023assessing} tried another 12 datasets. They found that although ChatGPT defeated preceding LLMs, it was still far from perfection. \citet{wang2023robustness} assessed ChatGPT on outliers, with similar results. Besides generating bad responses, the unsuccessful handling of such inputs may sometimes cause  more serious consequences, for example, data leakage or Denial of Service attacks. \citet{peng2023security} assessed whether ChatGPT would suffer from SQL injection if it served as a text-to-SQL interface~\citep{li2023can}, while the answer was positive.

\noindent \textbf{Reliability} requires the generated text to be faithful. Neural text generators hallucinate~\citep{ji2023survey} and, thus, produce unfaithful texts. ChatGPT/GPT-4 is no exception~\citep{bang2023multitask}. \citet{zhuo2023red} assessed ChatGPT on fact-based question-answering datasets and concluded that it was not improved compared to its predecessors. This makes ChatGPT unreliable in tasks where faithfulness is vital. For example, it might make up references when writing scientific articles~\citep{athaluri2023exploring} and make up legal cases when serving in the legal domain~\citep{deroy2023ready}. 

\noindent \textbf{Toxicity} asks models not to generate harmful, offensive and pornographic content. The GPT models were designed to normally refuse to generate toxic content. Nevertheless,~\citet{derner2023safeguards} found that through role-playing ChatGPT still produced offensive content. A quantitative study by~\citet{zhuo2023red} showed that, by feeding toxic prompting, only 0.5\% of the ChatGPT responses were toxic. Nonetheless, this ability was not very robust. In line with~\citet{derner2023safeguards}, they also found that it was susceptible to prompt injections achieved by role-playing. 

\noindent \textbf{Domain-specific concerns} were raised in cyberscurity~\citep{sebastian2023chatgpt}, marketing~\citep{rivas2023marketing}, politics~\citep{rozado2023political,motoki2023more}, legal~\citep{deroy2023ready}, education~\citep{sallam2023chatgpt,lee2023rise,kasneci2023chatgpt}, academic writing~\citep{athaluri2023exploring}, recommendation~\citep{zhang2023chatgpt} and tourism~\citep{carvalho2023chatgpt}. These concerns highlighted the potential misuse of the GPTs, the spread of misinformation, and the erosion of privacy.

\section{Discussion}

\subsection{Comparing GPT versus Humans}

For NLP tasks with rich training resources, the GPTs may not perform as well as expert models (see Tables~\ref{tab:ling_models} and \ref{tab:multiling_models} in Appendix~\ref{sec:visualisation}). There is also a certain distance from humans (see Tables~\ref{tab:ling_humans} and \ref{tab:multiling_humans}). However, when it comes to scientific knowledge, extensive multidisciplinary testing has showcased their superiority compared to earlier models (see Table~\ref{tab:science_models}). Notably, in computer science and law exams, GPT-4 has achieved a level of accuracy that closely matches or even surpasses the average human performance (see Table~\ref{tab:science_humans}). However, when we are quoted a study that shows an AI system performing at a similar level to humans, these are typically based on average performance over many cases. This average hides the fact that the AI is typically outperforming on some specialist knowledge that is difficult for humans, and underperforming on some examples that are relatively easy for humans. The same has been shown in computer vision~\citep{russakovsky2015imagenet}. Thus, human-level average accuracy achieved by AI is not equivalent to human performance and intelligence.

The GPTs may exhibit instances of hallucinations and false information~\citep{cabrera2023zeno,bang2023multitask}. This may be attributed to the fact that the ``next word prediction''-based pre-training only taught GPTs ``what is right'' while neglecting ``what is wrong''. To learn to generate a correct last word, e.g., ``bird'' after the context ``if an animal has wings and can fly, it is likely a'', GPTs have to learn logic, commonsense, linguistics, and science. However, without learning ``if an animal has wings and can fly, it is a penguin'' is wrong, GPTs may struggle to distinguish penguins from other birds by flight ability and yield penguins' flying hallucinations, because GPTs likely have learned ``penguins are birds, having wings'' from corpora. In reality, incorrect examples are far more than the negations we can see from corpora. False information may be exhibited in ambiguous cases if GPTs do not know the boundary between positive and negative examples. In contrast, humans learn knowledge from both right and wrong applications to sidestep obvious fallacies~(NASEM,~\citeyear{national2018people}).

The inference process of GPT models is also dissimilar to humans. Humans are known to have two types of reasoning: ``thinking fast'' and ``thinking slow''~\citep{daniel2000thinking}. However, GPT models seem to only have the thinking fast-like inference mechanism, e.g., they do a feedforward propagation without thinking twice~\citep{bubeck2023sparks}. Humans in contrast can spend longer deliberating for hard questions. This requires a longer process of iterative inference, not necessarily dictated by a fixed number of iterations, but instead iterating until the system achieves a satisfactory state~\citep{VANBERGEN2020176}. Nevertheless, it is clear that GPT models can do a certain type of reasoning. Especially, it has been shown that adding ``Let's think step by step'' to a prompt can allow GPT to use its own output as a sort of scratchpad to help it chain together multiple steps to arrive at a solution~\citep{bubeck2023sparks}. In this way GPT is simulating ``thinking slow'', however, it is limited to ``linear'' sequences of thought and has severe limitations in tasks that require planning (ibid.); it does not backtrack to try other possible alternatives. Such backtracking would require some sort of short-term memory or workspace to remember what has been tried and what is yet to be tried. These are well-known shortcomings of neural models~\citep{minsky1991logical}.

\citet{bubeck2023sparks}  noted the context-dependence of GPT's mathematical knowledge; ``changes in the wording of the question can alter the knowledge that the model displays.'' The general phenomenon of sensitivity to input phrasing is well known from other language models also~\citep{mao2023biases}.~\citet{lai2023chatgpt} found a huge drop in commonsense knowledge when testing in languages other than English. This also explains why assessments by different papers can arrive at contradictory conclusions about the GPT's knowledge. It demonstrates that the internal representation in GPT suffers from severe entanglement~\citep{6472238}. While more extensive training can cause it to give correct responses in more contexts, it cannot change the fundamental fact that knowledge is not disentangled from the text contexts in which it appears; therefore there will always be the possibility that a change to the text of a question, that preserves the semantics, will elicit a factually incorrect response. Fixing this issue may require a hybrid system with separate facts. A commonsense-based neurosymbolic AI framework, such as the one proposed by~\citet{cambria-etal-2022-senticnet} for sentiment analysis, moreover, can help increase the explainability of the reasoning processes required for decision-making, which is crucial for sensitive applications involving ethics, privacy and health.

\begin{figure*}[tbh!]
\centering
\includegraphics[width=\textwidth]{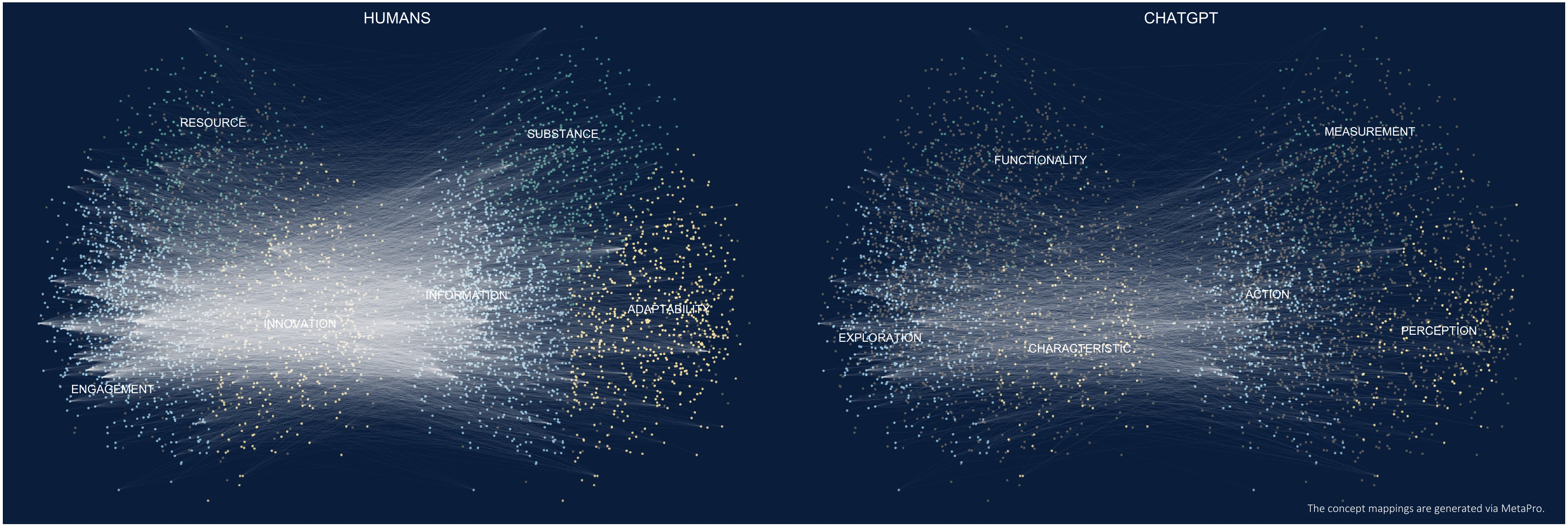} 
\caption{The concept mapping patterns between humans (left) and ChatGPT (right) from~\citet{mao2024comparative}. Each cluster on the left represents target concepts, while on the right, the cluster represents source concepts. Bright and grey dots denote activated and unactivated concepts, respectively. The capitalized terms represent key activated concepts within a cluster.}\label{fig:all_concept_mappings}
\end{figure*}

\subsection{Evaluation} \label{sec:eval}

Concerns about the validity of NLP evaluations in general~\citep{belz2023missing} are applicable to this paper also. The following points are worth noting: 

\noindent \textbf{1)}~ChatGPT as a commercial product is updated periodically~\citep{tu2023chatlog}. This means the ``flaws'' identified in early studies are fixed in later versions and different studies about the same task are not fully comparable. Additionally, since the repair may be done by including datasets from previous assessments in fine-tuning, the data leakage may make the assessment unfair~\citep{min-etal-2022-rethinking, openai2023gpt4}. This has been well-evidenced in~\citet{aiyappa2023trust} and~\citet{dewynter2023evaluation}. 

\noindent \textbf{2)}~The design of prompts highly influences the results, leading to biased comparisons. Take MT as an example, by seeking better prompts,~\citet{hendy2023good} and~\citet{jiao2023chatgpt} had very different conclusions compared to~\citet{bang2023multitask}. In future studies, it is necessary to ensure transparency in prompt design and facilitate a fair comparison between LLMs and baselines. 

\noindent \textbf{3)}~The factors that matter in previous NLP evaluations are still valid, which may include the choice of evaluation corpora and metrics, the design of human evaluations, the task formulation, and so on. For instance,~\citet{martinez2023re} argued that OpenAI's assessment of GPT-4~\citep{openai2023gpt4} on the Uniform Bar Exam is misleading because they incorrectly included test-takers who re-took the exam in their comparison. \citet{zhang2023exploring} misconcluded that GPT-4 had solved all MIT math and computer science curricula because they improperly used GPT-4 as the judge. 

\noindent \textbf{4)}~The assessments of some key abilities, e.g., creativity and logical reasoning, lack either objective criteria or large-scale benchmarks. Consequently, the evaluation can only capture a portion of the overall capabilities of GPT. People seem to commonly believe that AI will cause mass unemployment~\citep{hatzius2023potentially}. However, in these fields, the evaluation is also very limited. It would be expected to see what kind of leading results the GPT models achieved compared with humans in the field of work that can be replaced by AI.

\subsection{Ethics}

Several works have found that human perception about the reliability of ChatGPT's output can be misled by its seemingly scientific language style~\citep{gudibande2023false,dahlkemper2023physics}. Given AI-generated content, it would be necessary to allow users to retrieve source references that were generated by humans to improve response reliability. Caution should be also exercised when employing AI-generated data for training new AI models, as this practice carries the risk of introducing irreparable flaws into the resultant models~\citep{shumailov2023curse}. \citet{mao2024comparative} comparatively analyzed the cognitive patterns of ChatGPT and humans via metaphoric concept mappings. They used MetaPro~\citep{mao2023metaproonline} to extract target and source concepts from parallel answers provided by ChatGPT and humans to the same questions. Then, they plotted the concept distribution for each subject. In Figure~\ref{fig:all_concept_mappings}, compared to the diverse activated concepts from humans, ChatGPT primarily activated the blue concepts, leaving many concepts greyed out (unactivated) in other areas, indicating concepts and concept mappings that were not utilized in its generated text. It shows the concept preference of ChatGPT, suggesting potential cognition biases from its generated text. Consequently, caution is advised when utilizing AI-generated text for training subsequent AI models, as it may propagate biases to these models.

Even if the Reinforcement Learning from Human Feedback~\citep[RLHF,][]{openai2023chatgpt} can somewhat mitigate biased, inaccurate, and toxic responses, and enhance human-centric output preferences, RLHF may be misled by human-biased feedback, e.g., system gaming~\citep{leike2017learning}, positive reward cycles~\citep{ho2015teaching}, and tangled social norms and cultural context~\citep{liu2023perspectives}. There are also concerns regarding the non-transparency of the training set of ChatGPT, which could potentially lead to data leakage, including trade secrets or personal privacy, as user inputs might be utilized for fine-tuning~\citep{min-etal-2022-rethinking}. These limitations appearing in the training and inferring processes may raise additional biases and ethical concerns.

\section{Conclusion and Recommendations}
In this work, we reviewed the latest assessments of ChatGPT and GPT-4, including their language and reasoning ability, scientific knowledge, and relevant ethical considerations. The former assessments showed that the GPTs have demonstrated strong capability in language understanding and generation, as well as general scientific knowledge. On the other hand, we also observed that the GPTs fall behind expert systems in many conventional NLP tasks; Their multi-step reasoning skills could benefit from further development; Ethical considerations in fairness, robustness, reliability, toxicity, and different application domains remain; The current evaluation tasks can be further improved with better transparency in GPT training corpora and evaluation methodology, as well as broader testing domains. In addition to implementing effective measures to address the aforementioned issues, the following aspects are also recommended:

\noindent \textbf{Task-agnostic evaluation is desirable.} Our survey indicates that many evaluations of ChatGPT and GPT-4 have primarily relied on benchmark data or task-specific inquiries. However, this method has limitations, e.g., the risk of data contamination during the pre-training phase, the variability of LLM performance across different prompts and task formulations, and the narrow scope of the evaluation.

To address these limitations, task-agnostic evaluation approaches offer a promising solution. In psychology, a common practice for assessing mental health involves using concept mappings derived from indirect questions and answers that reflect emotional states and cognition. These indirect tests, such as word-association tests, thematic apperception tests, and the Rorschach test~\citep{rapaport1946diagnostic}, elicit responses that reflect the subconscious mind, which is believed to influence the majority of brain activities~\citep{pally2007predicting} and human behaviors~\citep{pradeep2010buying}. Similarly, for non-open-source LLMs, such as ChatGPT and GPT-4, researchers can develop indirect tasks to detect potential issues by considering LLMs as subjects. For example, cognitive bias analysis can be conducted by comparing concept mappings derived from human-generated and ChatGPT-generated text, particularly in response to questions from other domains~\citep{mao2024comparative}. This comparative analysis can reveal discrepancies that may indicate biases or limitations in the model's understanding. For open-source neural networks, neuro-activation variation analysis can also be employed to identify potential deficiencies in pre-trained language models~\citep{chen2021badpre}.

\noindent \textbf{Fundamental research is still valuable.} The landscape of NLP has been largely influenced by the emergence of ChatGPT-like LLMs. Researchers are obsessed with using them to process high-level application tasks. Comparatively, conventional computational linguistics tasks, e.g., syntactic and semantic processing~\citep{zhang2023syntactic,mao2024semantic}, have taken a backseat in recent times. This shift is driven by the belief that the direct utility of these foundational techniques is less pronounced in enhancing model performance on high-level tasks. For instance, semantic knowledge can now be acquired through extensive pre-training, obviating the need for explicitly distinguishing parts of speech or word senses.

However, the significance of computational linguistics research extends beyond the development of NLP applications. It also offers insights into how humans perceive and use language. Fundamental research in computational linguistics remains valuable, as in many resource-rich areas, ``expert models'' still lead by some distance, compared to the GPTs. However, the scope of the fundamental tasks must be broadened to consider more than simple classification based on input text, encompassing, e.g., broader narrative~\citep{cambria2014jumping} or cognitive~\citep{ge2023survey} understanding. The value of studying fundamental tasks is also manifested in the spirit of task decomposition -- one of the important pillars of AI~\citep{cambria2023seven}. Failing to break down a complex task into its component subtasks effectively requires the model to implicitly address numerous subtasks for which it has not been specifically trained. This approach undermines the establishment of trustworthiness and accountability.

\noindent \textbf{AI-generated content should be regulated.} To keep pace with the rapid development of generative AI, regulatory frameworks must also evolve swiftly to prevent the misuse of AI and AI-generated content. The success of current LLMs largely stems from their exposure to diverse human-generated content from various sources and perspectives. This diversity enables models like GPTs to learn from a wide range of ideas and generate output that aligns with the expectations of the majority.

However, the speed at which machines can generate content far exceeds that of humans. As shown in Figure~\ref{fig:all_concept_mappings}, humans presented greater diversity in metaphorical concepts and concept mappings compared to ChatGPT, reflecting the richness of human thought. Without proper controls, a large volume of AI-generated content may blend with human-generated content, potentially diluting the diversity of human thought. This scenario could lead to individuals perceiving opinions from a single major source -- machines. Moreover, the misuse of AI-generated content for training could introduce biases into subsequent AI~\citep{mao2024comparative}. Given the aforementioned ethical considerations and the limitations of current evaluation methods, it is necessary to properly regulate AI-generated content before knowing how to use them well. Otherwise, as~\citet{cambria2023seven} argued: ``AI could very much end up being like plastic: a great invention that made our life easier about a century ago but is now threatening our own existence.''

\noindent \textbf{The Future of LLMs.} The major advances in recent LLMs have largely come from scaling up. At this time so much text training data is already being used that the potential for scaling up further is limited. Recently interest has moved to multimodal models, for example, those that use images as well as text \cite{yang2023dawn,mckinzie2024mm1}, and other modalities \cite{wang2023large}. This opens a much wider range of human tasks that these models could potentially be applied to. Some of the major shortcomings of LLMs such as hallucination or inaccuracy, and inability to do ``thinking slow'' reasoning, seem unlikely to be solved at a foundational level in the near future. Instead, we see LLMs using external resources to compensate for their deficiencies, e.g. they can fact-check using provided resources, or the Internet, or they can check calculations with a calculator, or check code with an interpreter. Furthermore, multimodal models can potentially be proactive in going to websites and performing tasks using the graphical interface as a human would. LLMs would be conceptualized as the kernel process of a new operating system, facilitating interactions with various resources to accomplish tasks.

\section*{Acknowledgments}
This research/project is supported by the Ministry of Education, Singapore under its MOE Academic Research Fund Tier 2 (STEM RIE2025 Award MOE-T2EP20123-0005). Guanyi Chen is supported by the Hubei Provincial Key Laboratory of Artificial Intelligence and Smart Learning and the National Language Resources Monitoring and Research Center for Network Media of Central China Normal University in Wuhan, China.

\nocite{*}
\section{Bibliographical References}\label{sec:reference}

\bibliographystyle{lrec-coling2024-natbib}
\bibliography{reference}

%\clearpage
\appendix

\section{ChatGPT and GPT-4 benchmark}
\label{sec:visualisation}

We visualize the performance gaps between the surveyed GPT models and baselines on different tasks in Tables~\ref{tab:ling_models}-\ref{tab:science_humans}. Task abbreviations and meanings can be viewed in Table~\ref{tab:abbr}. The visualization illustrates different levels of disparity in performance between ChatGPT or GPT-4 and the most robust baseline presented in an evaluation paper. Darker colors represent greater disparities, while lighter colors indicate smaller differences. The degree of discrepancy is measured by $g/b - 1$, where $g$ is the average score of a GPT model on the major metric of an evaluation task; $b$ is the average score of a baseline on the same major metric. The color red indicates instances where the GPT models outperformed the baselines, while blue indicates cases where the GPT models lagged behind. The teal color represents the gaps between the GPT models and the ground truth, while gray indicates a lack of comparison. The summary of the surveyed assessment papers can be viewed in Tables~\ref{tab:summary} and~\ref{tab:summary_findings}.

\begin{table*}
\centering
\scriptsize
\begin{tabular}{ll|ll} 
\toprule
\multicolumn{1}{c}{\textbf{Abbr.}} & \multicolumn{1}{c|}{\textbf{Meaning}} & \multicolumn{1}{c}{\textbf{Abbr.}} & \multicolumn{1}{c}{\textbf{Meaning}} \\ 
\hline
ACI & Actual Causality Inferring  & Natu.S  & Natural Science  \\
Asp.E & Aspect Extraction & NER & Named Entity Recognition \\
Asp.PD  & Aspect Polarity Detection & NLI & Natural Language Inference \\
Caus.D  & Cause Discovery & Offe.D  & Offensiveness Detection  \\
Caus.R  & Causal Reasoning  & OKEP  & Ophthalmic Knowledge Assessment Program  \\
CBQA  & Complex Boolean Question Answering  & Opin.E  & Opinion Extraction \\
CCSE  & Chinese Civil Service Examination & Overa.  & Overall  \\
\bottomrule
\end{tabular}
\caption{Task abbreviations and meanings.}\label{tab:abbr}
\end{table*}

\begin{table*}[!bh]
  \centering
  \scriptsize
  \begin{tabular}{llp{0.04cm}p{0.04cm}p{0.04cm}p{0.04cm}p{0.04cm}p{0.04cm}p{0.04cm}p{0.04cm}p{0.04cm}p{0.04cm}p{0.04cm}p{0.04cm}p{0.04cm}p{0.04cm}p{0.04cm}p{0.04cm}p{0.04cm}p{0.04cm}|p{0.04cm}p{0.04cm}p{0.04cm}p{0.04cm}p{0.04cm}p{0.04cm}p{0.04cm}p{0.04cm}p{0.04cm}}
  &  & \multicolumn{18}{c}{\bf ChatGPT}  & \multicolumn{8}{c}{\bf GPT-4}  \\
  &  & \begin{sideways}\citet{cabrera2023zeno}\end{sideways} & \begin{sideways}\citet{amin2023wide}\end{sideways} & \begin{sideways}\citet{kocon2023chatgpt}\end{sideways} & \begin{sideways}\citet{liu2023evaluating}\end{sideways} & \begin{sideways}\citet{lai2023chatgpt}\end{sideways} & \begin{sideways}\citet{qin2023chatgpt}\end{sideways} & \begin{sideways}\citet{wei2023zero}\end{sideways} & \begin{sideways}\citet{hendy2023good}\end{sideways} & \begin{sideways}\citet{bang2023multitask}\end{sideways} & \begin{sideways}\citet{gao2023chatgpt}\end{sideways} & \begin{sideways}\citet{kiciman2023causal}\end{sideways} & \begin{sideways}\citet{wang2023chatgpt}\end{sideways} & \begin{sideways}\citet{kocmi2023large}\end{sideways} & \begin{sideways}\citet{jiao2023chatgpt}\end{sideways} & \begin{sideways}\citet{karpinska2023large}\end{sideways} & \begin{sideways}\citet{zhang2023extractive}\end{sideways} & \begin{sideways}\citet{sun2023chatgpt}\end{sideways} & \begin{sideways}\citet{liu2023were}\end{sideways} & \begin{sideways}\citet{amin2023wide}\end{sideways} & \begin{sideways}\citet{liu2023evaluating}\end{sideways} & \begin{sideways}\citet{gao2023chatgpt}\end{sideways} & \begin{sideways}\citet{kiciman2023causal}\end{sideways} & \begin{sideways}\citet{liu2023gpteval}\end{sideways} & \begin{sideways}\citet{jiao2023chatgpt}\end{sideways} & \begin{sideways}\citet{sun2023chatgpt}\end{sideways} & \begin{sideways}\citet{bubeck2023sparks}\end{sideways} & \begin{sideways}\citet{liu2023were}\end{sideways}  \\
\end{tabular}
  \caption{ChatGPT and GPT-4 performance on linguistic and reasoning tasks, compared to baseline models. Supv denotes supervised models. The column names denote the type of the strongest baseline, introduced in an evaluation paper. The row names indicate evaluation tasks.}\label{tab:ling_models}
  \end{table*}

\begin{table*}[!thb]
\centering
\scriptsize
\begin{tabular}{llllllllll|lllll}
  & & \multicolumn{8}{c}{\textbf{ChatGPT}}  & \multicolumn{5}{c}{\textbf{GPT-4}}  \\
  & & \begin{sideways}\citet{hendy2023good}\end{sideways} & \begin{sideways}\citet{bang2023multitask}\end{sideways} & \begin{sideways}\citet{gao2023chatgpt}\end{sideways} & \begin{sideways}\citet{kiciman2023causal}\end{sideways} & \begin{sideways}\citet{kiciman2023causal}\end{sideways} & \begin{sideways}\citet{liu2023evaluating}\end{sideways} & \begin{sideways}\citet{liu2023were}\end{sideways} & \begin{sideways}\citet{gilardi2023chatgpt}\end{sideways} & \begin{sideways}\citet{liu2023evaluating}\end{sideways} & \begin{sideways}\citet{gao2023chatgpt}\end{sideways} & \begin{sideways}\citet{kiciman2023causal}\end{sideways} & \begin{sideways}\citet{kiciman2023causal}\end{sideways} & \begin{sideways}\citet{liu2023were}\end{sideways}  \\
\multicolumn{2}{l}{Task} & \begin{sideways}Human avg\end{sideways} & \begin{sideways}Ground truth\end{sideways}  & \begin{sideways}Human avg\end{sideways}  & \begin{sideways}Human avg\end{sideways} & \begin{sideways}Ground true\end{sideways} & \begin{sideways}Human avg\end{sideways} & \begin{sideways}Human avg\end{sideways} & \begin{sideways}Human avg\end{sideways}  & \begin{sideways}Human avg\end{sideways} & \begin{sideways}Human avg\end{sideways}  & \begin{sideways}Human avg\end{sideways} & \begin{sideways}Ground true\end{sideways} & \begin{sideways}Human avg\end{sideways}  \\
Gene. & Mach.T  & \cellcolor{blue!3.69415232} & \cellcolor{gray!30} & \cellcolor{gray!30}  & \cellcolor{gray!30} & \cellcolor{gray!30} & \cellcolor{gray!30} & \cellcolor{gray!30} & \cellcolor{gray!30}  & \cellcolor{gray!30} & \cellcolor{gray!30}  & \cellcolor{gray!30} & \cellcolor{gray!30} & \cellcolor{gray!30}  \\
\end{tabular}
\caption{ChatGPT and GPT-4 performance on linguistic and reasoning tasks, compared to humans or ground truth.}\label{tab:ling_humans}
\end{table*}

\begin{table}[!tbh]
 \centering
 \scriptsize
 \begin{tabular}{p{0.3cm}p{0.66cm}p{0.05cm}p{0.05cm}p{0.05cm}p{0.05cm}p{0.05cm}|p{0.05cm}p{0.05cm}p{0.05cm}|p{0.05cm}p{0.66cm}p{0.05cm}}
 \multicolumn{1}{c}{} & \multicolumn{1}{c}{} & \multicolumn{5}{c}{\bf ChatGPT} & \multicolumn{3}{c}{\bf GPT-4} & \multicolumn{3}{c}{\bf ChatGPT} \\
 & & \begin{sideways}Xu et al. (\citeyear{SuperCLUE})\end{sideways} & \begin{sideways}\citet{huang2023c}\end{sideways} & \begin{sideways}\citet{lai2023chatgpt}\end{sideways} & \begin{sideways}\citet{wei2023zero}\end{sideways} & \begin{sideways}\citet{wang2023cross}\end{sideways} & \begin{sideways}Xu et al. (\citeyear{SuperCLUE})\end{sideways} & \begin{sideways}\citet{huang2023c}\end{sideways} & \begin{sideways}\citet{wang2023cross}\end{sideways} & & & \begin{sideways}\citet{lai2023chatgpt}\end{sideways} \\
\multicolumn{2}{l}{Task} & \begin{sideways}LLM\end{sideways} & \begin{sideways}LLM\end{sideways} & \begin{sideways}Supv\end{sideways} & \begin{sideways}Supv\end{sideways} & \begin{sideways}SOTA\end{sideways} & \begin{sideways}LLM\end{sideways} & \begin{sideways}LLM\end{sideways} & \begin{sideways}SOTA\end{sideways} & \begin{sideways}\end{sideways} & \begin{sideways}\end{sideways} & \begin{sideways}Supv\end{sideways} \\
 en2zh & Summ. & \cellcolor{gray!30} & \cellcolor{gray!30} & \cellcolor{gray!30} & \cellcolor{gray!30} & \cellcolor{blue!18.35622928} & \cellcolor{gray!30} & \cellcolor{gray!30} & \cellcolor{blue!24.01705353} & jp & Com.R & \cellcolor{blue!22.83609576} \\
 en2de & Summ. & \cellcolor{gray!30} & \cellcolor{gray!30} & \cellcolor{gray!30} & \cellcolor{gray!30} & \cellcolor{blue!14.72733588} & \cellcolor{gray!30} & \cellcolor{gray!30} & \cellcolor{blue!10.73289337} & vi & PoS & \cellcolor{red!17.39130435} \\
 ru & PoS & \cellcolor{gray!30} & \cellcolor{gray!30} & \cellcolor{red!5.408515535} & \cellcolor{gray!30} & \cellcolor{gray!30} & \cellcolor{gray!30} & \cellcolor{gray!30} & \cellcolor{gray!30} & vi & NLI & \cellcolor{blue!36.02771363} \\
\end{tabular}
 \caption{ChatGPT and GPT-4 performance on multi-lingual tasks, compared to baseline models. Languages are chunked by high (en2zh-vi), medium (tr-hi), low (bn-kn), and extremely low (sw-as) language resources, respectively, categorized by~\citet{lai2023chatgpt}.}\label{tab:multiling_models}
 \end{table}

\begin{table}[!tbh]
\centering
\scriptsize
\begin{tabular}{llll|l}
 & & \multicolumn{2}{c}{\textbf{ChatGPT}} & \multicolumn{1}{c}{\textbf{G4}} \\
 & & \begin{sideways}Xu et al. (\citeyear{SuperCLUE})\end{sideways} & \begin{sideways}\citet{bang2023multitask}\end{sideways} & \begin{sideways}Xu et al. (\citeyear{SuperCLUE})\end{sideways} \\
\multicolumn{2}{l}{Task} & \begin{sideways}Human\end{sideways} & \begin{sideways}Ground truth\end{sideways} & \begin{sideways}Human\end{sideways} \\
zh & Ling.A & \cellcolor{blue!13.26530612} & \cellcolor{gray!30} & \cellcolor{blue!8.163265306} \\
zh & Prof.A & \cellcolor{blue!37.89473684} & \cellcolor{gray!30} & \cellcolor{blue!28.42105263} \\
zh & Overa. & \cellcolor{blue!31.41968912} & \cellcolor{gray!30} & \cellcolor{blue!20.5492228} \\
id & Sent.A & \cellcolor{gray!30} & \cellcolor{teal!20} & \cellcolor{gray!30} \\
id & LID & \cellcolor{gray!30} & \cellcolor{teal!0} & \cellcolor{gray!30} \\
jw & Sent.A & \cellcolor{gray!30} & \cellcolor{teal!22} & \cellcolor{gray!30} \\
jw & LID & \cellcolor{gray!30} & \cellcolor{teal!100} & \cellcolor{gray!30} \\
bug & Sent.A & \cellcolor{gray!30} & \cellcolor{teal!44} & \cellcolor{gray!30} \\
bug & LID & \cellcolor{gray!30} & \cellcolor{teal!88} & \cellcolor{gray!30} 
\end{tabular}
\caption{ChatGPT and GPT-4 performance on multi-lingual tasks, compared to humans or ground truth.}\label{tab:multiling_humans}
\end{table}

\begin{table}[!tbh]
 \centering
 \scriptsize
 \begin{tabular}{llp{0.06cm}p{0.06cm}p{0.06cm}p{0.06cm}p{0.06cm}p{0.06cm}p{0.06cm}|p{0.06cm}p{0.06cm}}
 & & \multicolumn{7}{c}{\textbf{ChatGPT}} & \multicolumn{2}{c}{\textbf{GPT-4}} \\
 & & \begin{sideways}\citet{gilson2023does}\end{sideways} & \begin{sideways}\citet{liu2023evaluating}\end{sideways} & \begin{sideways}\citet{qin2023chatgpt}\end{sideways} & \begin{sideways}\citet{li2023can}\end{sideways} & \begin{sideways}\citet{xie2023wall}\end{sideways} & \begin{sideways}\citet{kung2023performance}\end{sideways} & \begin{sideways}\citet{bubeck2023sparks}\end{sideways} & \begin{sideways}\citet{liu2023evaluating}\end{sideways} & \begin{sideways}\citet{bubeck2023sparks}\end{sideways} \\
 \multicolumn{2}{l}{Task} & \begin{sideways}Old GPT\end{sideways} & \begin{sideways}Supv\end{sideways} & \begin{sideways}Supv\end{sideways} & \begin{sideways}SOTA\end{sideways} & \begin{sideways}SOTA\end{sideways} & \begin{sideways}Expert LLM\end{sideways} & \begin{sideways}SOTA\end{sideways} & \begin{sideways}Supv\end{sideways} & \begin{sideways}SOTA\end{sideways} \\
 Form.S & T2SQL & \cellcolor{gray!30} & \cellcolor{gray!30} & \cellcolor{gray!30} & \cellcolor{red!9.898546751} & \cellcolor{gray!30} & \cellcolor{gray!30} & \cellcolor{gray!30} & \cellcolor{gray!30} & \cellcolor{gray!30} \\
 Form.S & Coding & \cellcolor{gray!30} & \cellcolor{gray!30} & \cellcolor{gray!30} & \cellcolor{gray!30} & \cellcolor{gray!30} & \cellcolor{gray!30} & \cellcolor{red!66.66666667} & \cellcolor{gray!30} & \cellcolor{red!100} \\
 Form.S & LeetCo. & \cellcolor{gray!30} & \cellcolor{gray!30} & \cellcolor{gray!30} & \cellcolor{gray!30} & \cellcolor{gray!30} & \cellcolor{gray!30} & \cellcolor{red!46.15384615} & \cellcolor{gray!30} & \cellcolor{red!100} \\
 Form.S & Math & \cellcolor{gray!30} & \cellcolor{gray!30} & \cellcolor{red!2.634934127} & \cellcolor{gray!30} & \cellcolor{gray!30} & \cellcolor{gray!30} & \cellcolor{blue!14.03834261} & \cellcolor{gray!30} & \cellcolor{red!31.29251701} \\
\end{tabular}
 \caption{ChatGPT and GPT-4 performance on scientific knowledge, compared to baselines.}\label{tab:science_models}
 \end{table}

\begin{table*}[!thb]
\centering
\scriptsize
\begin{tabular}{llp{0.066cm}p{0.066cm}p{0.066cm}p{0.066cm}p{0.066cm}p{0.066cm}p{0.066cm}p{0.066cm}p{0.066cm}p{0.066cm}p{0.066cm}p{0.066cm}p{0.066cm}p{0.066cm}p{0.066cm}p{0.066cm}p{0.066cm}|p{0.066cm}p{0.066cm}p{0.066cm}p{0.066cm}}
 & & \multicolumn{17}{c}{\textbf{ChatGPT}} & \multicolumn{4}{c}{\textbf{GPT-4}} \\
 & & \begin{sideways}\citet{li2023can}\end{sideways} & \begin{sideways}\citet{bordt2023chatgpt}\end{sideways} & \begin{sideways}\citet{clark2023investigating}\end{sideways} & \begin{sideways}\citet{liu2023evaluating}\end{sideways} & \begin{sideways}\citet{antaki2023evaluating}\end{sideways} & \begin{sideways}\citet{hirosawa2023diagnostic}\end{sideways} & \begin{sideways}\citet{mehnen2023chatgpt}\end{sideways} & \begin{sideways}\citet{bubeck2023sparks}\end{sideways} & \begin{sideways}\citet{dahlkemper2023physics}\end{sideways} & \begin{sideways}\citet{frieder2023mathematical}\end{sideways} & \begin{sideways}\citet{kortemeyer2023could}\end{sideways} & \begin{sideways}\citet{choi2023chatgpt}\end{sideways} & \begin{sideways}\citet{geerling2023chatgpt}\end{sideways} & \begin{sideways}\citet{xie2023wall}\end{sideways} & \begin{sideways}\citet{gilson2023does}\end{sideways} & \begin{sideways}\citet{kung2023performance}\end{sideways} & \begin{sideways}\citet{antaki2023evaluating}\end{sideways} & \begin{sideways}\citet{bordt2023chatgpt}\end{sideways} & \begin{sideways}\citet{liu2023evaluating}\end{sideways} & \begin{sideways}\citet{mehnen2023chatgpt}\end{sideways} & \begin{sideways}\citet{bubeck2023sparks}\end{sideways} \\
\end{tabular}
\caption{ChatGPT and GPT-4 performance on scientific knowledge, compared to humans or ground truth.}\label{tab:science_humans}
\end{table*}

\begin{sidewaystable*}
\centering
\tiny
\begin{tabularx}{\textwidth}{llllll} 
\toprule
\multicolumn{1}{c}{\textbf{Reference}} & \multicolumn{1}{c}{\textbf{Evaluation task}} & \multicolumn{1}{c}{\textbf{Version}} & \multicolumn{1}{c}{\textbf{Date}} & \multicolumn{1}{c}{\textbf{Evaluation method}} & \multicolumn{1}{c}{\textbf{Evaluation method novelty}}\\ 
\hline
\citet{cabrera2023zeno} & Dialogue fluency & CG, G4 & 18/05 & Compare to zero-shot LLMs on benchmark data & Unified interface for open-source LLMs; Online browsing and analysis \\ 
\hline
\citet{bang2023multitask} & Summ., Mach.T, Sent.A, QA, Misi.D & CG & 28/02 & Compare to SOTA on benchmark data & Multitask, multilingual, multimodal evaluation \\ 
\hline
\citet{hendy2023good} & Machine translation & CG & 18/02 & Compare to SOTA and other LLMs on benchmark data & Evaluating by prompting strategies, languages, document-level and domain robustness \\ 
\hline
\citet{jiao2023chatgpt} & Machine translation & CG, G4 & 19/03 & Compare to commercial MT systems on benchmark data & Testing robustness of MT systems on domain biases and noise \\ 
\hline
\citet{karpinska2023large} & Machine translation & CG, G4 & 22/05 & Compare to commercial MT systems on literacy texts & Conducting large scale human evaluation by experts and error analysis \\ 
\hline
\citet{qin2023chatgpt} & NLP and reasoning tasks & CG & 15/02 & Compare to other LLMs and supervised models on benchmark data & A wide range of task-oriented reasoning tasks \\ 
\hline
\citet{zhang2023extractive} & Summarisation & CG & 09/04 & Compare to SOTA on benchmark data & Exploring several ways to enhance the performance  \\
\end{tabularx}
\caption{The summary of our surveyed works (Part I). The date denotes the available online date in 2023. CG denotes ChatGPT. G4 denotes GPT-4.}\label{tab:summary}
\end{sidewaystable*}

\begin{sidewaystable*}
\centering
\tiny
%%
\end{tabularx}
\caption{The summary of our surveyed works (Part II).}\label{tab:summary_findings}
\end{sidewaystable*}

\end{document}